\documentclass[review]{elsarticle}

\usepackage{hyperref}
\usepackage{amsmath}

\usepackage{graphicx}
\usepackage{subcaption}
\usepackage{float}
\usepackage{multirow}
\usepackage{natbib}
 \usepackage{xcolor}
\setcitestyle{numbers}
\setcitestyle{square}

\journal{ the journal}

\biboptions{authoryear}

\begin{document}

\begin{frontmatter}


\title{Detecting Fake News with\\
   Capsule Neural Networks}

\author{Mohammad Hadi Goldani}
\ead{goldani@aut.ac.ir}

\author{Saeedeh Momtazi\corref{cor2}}
\ead{momtazi@aut.ac.ir}

\author{Reza Safabakhsh}
\ead{safa@aut.ac.ir}

\address{Computer Engineering Department, Amirkabir University of Technology, Tehran, Iran}

\cortext[cor2]{Corresponding author}

\begin{abstract}
Fake news is dramatically increased in social media in recent years. This has prompted the need for effective fake news detection algorithms. Capsule neural networks have been successful in computer vision and are receiving attention for use in Natural Language Processing (NLP). This paper aims to use capsule neural networks in the fake news detection task. We use different embedding models for news items of different lengths. Static word embedding is used for short news items, whereas non-static word embeddings that allow incremental up-training and updating in the training phase are used for medium length or large news statements. Moreover, we apply different levels of n-grams for feature extraction. Our proposed architectures are evaluated on two recent well-known datasets in the field, namely  ISOT and LIAR. The results show encouraging performance, outperforming the state-of-the-art methods by  7.8\% on ISOT and  3.1\% on the validation set, and 1\% on the test set of the LIAR dataset.
\end{abstract}


\end{frontmatter}


\section{Introduction}
\label{sec:intro}
Flexibility and ease of access to social media have resulted in the use of online channels for news access by a great number of people. For example, nearly two-thirds of American adults have access to news by online channels \citep{zhang2019overview,dale2017nlp}.  \citet{newman2013social} also reported that social media and news consumption is significantly increased in Great Britain.

In comparison to traditional media, social networks have proved to be more beneficial, especially during a crisis, because of the ability to spread breaking news much faster \citep{bondielli2019survey}. All of the news, however, is not real and there is a possibility of changing and manipulating real information by people due to political, economic, or social motivations. This manipulated data leads to the creation of news that may not be completely true or may not be completely false \citep{granik2017fake}.
Therefore, there is misleading information on social media that has the potential to cause many problems in society. Such misinformation, called fake news, has a wide variety of types and formats. Fake advertisements, false political statements, satires, and rumors are examples of fake news \citep{zhang2019overview}.
This widespread of fake news that is even more than mainstream media \citep{ balmas2014fake} motivated many researchers and practitioners to focus on presenting effective automatic frameworks for detecting fake news \citep{ horne2017just}. Google has announced an online service called “Google News Initiative” to fight fake news \citep{google:2018}. This project will try to help readers for realizing fake news and reports \citep{google:2018an}.

Detecting fake news is a challenging task. A fake news detection model tries to predict intentionally misleading news based on analyzing the real and fake news that previously reviewed. Therefore, the availability of high-quality and large-size training data is an important issue.

The task of fake news detection can be a simple binary classification or, in a challenging setting, can be a fine-grained classification \citep{tin2018study}.  After 2017, when fake news datasets were introduced, researchers tried to increase the performance of their models using this data. Kaggle dataset, ISOT dataset, and LIAR dataset are some of the most well-known publicly available datasets \citep{meel2019fake}.

In this paper, we propose a new model based on capsule neural networks for detecting fake news. We propose architectures for detecting fake news in different lengths of news statements by using different varieties of word embedding and applying different levels of n-gram as feature extractors. We show these proposed models achieve better results in comparison to the state-of-the-art methods.

The rest of the paper is organized as follows: Section  \ref{sec:Section 2} reviews related work about fake news detection. Section \ref{sec:Section 3} presents the model proposed in this paper. The datasets used for fake news detection and evaluation metrics are introduced in Section \ref{sec:Section 4}. Section \ref{sec:Section 5} reports the experimental results, comparison with the baseline classification and discussion. Section \ref{sec:Section 6} summarizes the paper and concludes this work.

\section{Related work}
\label{sec:Section 2}
Fake news detection has been studied in several investigations. \citet{Conroy:2015} presented an overview of deception assessment approaches, including the major classes and the final goals of these approaches. They also investigated the problem using two approaches: (1) linguistic methods, in which the related language patterns were extracted and precisely analyzed from the news content for making decision about it, and (2) network approaches, in which the network parameters such as network queries and message metadata were deployed for decision making about new incoming news.   

\citet{Ruchansky:2017} proposed an automated fake news detector, called CSI that consists of three modules: Capture, Score, and Integrate, which predicts by taking advantage of three features related to the incoming news:  text, response, and source of it. The model includes three modules; the first one extracts the temporal representation of news articles, the second one represents and scores the behavior of the users, and the last module uses the outputs of the first two modules (i.e., the extracted representations of both users and articles) and use them for the classification. Their experiments demonstrated that CSI provides an improvement in terms of accuracy.

\citet{Tacchini:2017} introduced a new approach which tries to decide if a news is fake or not based on the users that interacted with and/or liked it. They proposed two classification methods. The first method deploys a logistic regression model and takes the user interaction into account as the features. The second one is a novel adaptation of the Boolean label crowdsourcing techniques. The experiments showed that both approaches achieved high accuracy and proved that considering the users who interact with the news is an important feature for making a decision about that news.

\citet{PerezRosas:2018} introduced two new datasets that are related to seven different domains, and instead of short statements containing fake news information, their datasets contain actual news excerpts. They deployed a linear support vector machine classifier and showed that linguistic features such as lexical, syntactic, and semantic level features are beneficial to distinguish between fake and genuine news. The results showed that the performance of the developed system is comparable to that of humans in this area.

\citet{Wang:2017} provided a novel dataset, called LIAR, consisting of 12,836 labeled short statements. The instances in this dataset are chosen from more natural contexts such as Facebook posts, tweets, political debates, etc. They proposed neural network architecture for taking advantage of text and meta-data together. The model consists of a Convolutional Neural Network (CNN) for feature extraction from the text and a Bi-directional Long Short Term Memory (BiLSTM) network for feature extraction from the meta-data and feeds the concatenation of these two features into a fully connected softmax layer for making the final decision about the related news. They showed that the combination of metadata with text leads to significant improvements in terms of accuracy.

\citet{long2017fake} proved that incorporating speaker profiles into an attention-based LSTM model can improve the performance of a fake news detector. They claim speaker profiles can contribute to the model in two different ways. First, including them in the attention model. Second, considering them as additional input data. They used party affiliation, speaker location, title, and credit history as speaker profiles, and they show this metadata can increase the accuracy of the classifier on the LIAR dataset.

\citet{ahmed:2017} presented a new dataset for fake news detection, called ISOT. This dataset was entirely collected from real-world sources. They used n-gram models and six machine learning techniques for fake news detection on the ISOT dataset. They achieved the best performance by using  TF-IDF as the feature extractor and linear support vector machine as the classifier. 

\citet{Wang:2018} proposed an end-to-end framework called event adversarial neural network, which is able to extract event-invariant multi-modal features. This model has three main components: the multi-modal feature extractor, the fake news detector, and the event discriminator. The first component uses CNN as its core module. For the second component, a fully connected layer with softmax activation is deployed to predict if the news is fake or not. As the last component, two fully connected layers are used, which aims at classifying the news into one of K events based on the first component representations.

\citet{Tschiatschek:2018} developed a tractable Bayesian algorithm called Detective, which provides a balance between selecting news that directly maximizes the objective value and selecting news that aids toward learning user's flagging accuracy. They claim the primary goal of their works is to minimize the spread of false information and to reduce the number of users who have seen the fake news before it becomes blocked. Their experiments show that Detective is very competitive against the fictitious algorithm OPT, an algorithm that knows the true users’ parameters, and is robust in applying flags even in a setting where the majority of users are adversarial.

\section{Capsule networks for fake news detection}
\label{sec:Section 3}

In this section, we first introduce different variations of word embedding models. Then, we proposed two capsule neural network models according to the length of the news statements that incorporate different word embedding models for fake news detection.

\subsection{Different variations of word embedding models}
\label{sec:non-static}
Dense word representation can capture syntactic or semantic information from words. When word representations are demonstrated in low dimensional space, they are called word embedding. In these representations, words with similar meanings are in close position in the vector space.

In 2013, \citet{mikolov2013efficient} proposed word2vec, which is a group of highly efficient computational models for learning word embeddings from raw text. These models are created by training neural networks with two-layers trained by a large volume of text. These models can produce vector representations for every word with several hundred dimensions in a vector space.  In this space, words with similar meanings are mapped to close coordinates.

There are some pre-trained word2vec vectors like 'Google News' that was trained on 100 billion words from Google news. One of the popular methods to improve text processing performance is using these pre-trained vectors for initializing word vectors, especially in the absence of a large supervised training set. These distributed vectors can be fed into deep neural networks and used for any text classification task \citep{kim2014convolutional}. These pre-trained embeddings, however, can further be enhanced.

\citet{kim2014convolutional} applied different learning settings for vector representation of words via word2vec for the first time and showed their superiority compared to the regular pre-trained embeddings when they are used within a CNN model. These settings are as follow:
\begin{itemize}
\item \textbf{Static word2vec model:} in this model, pre-trained vectors are used as input to the neural network architecture, these vectors are kept static during training, and only the other parameters are learned.
\item \textbf{Non-static word2vec model:} this model uses the pre-trained vectors at the initialization of learning, but during the training phase, these vectors are fine-tuned for each task using the training data of the target task.
\item \textbf{Multichannel word2vec model:} the model uses two sets of static and non-static word2vec vectors, and a part of vectors fine-tune during training.
\end{itemize}

\subsection{Proposed model}

Although different models based on deep neural networks have been proposed for fake news detection, there is still a great need for further improvements in this task. In the current research, we aim at using capsule neural networks to enhance the accuracy of fake news identification systems.

The capsule neural network was introduced by \citet{ sabour:2017} for the first time in the paper called ``Dynamic Routing Between Capsules''. In this paper, they showed that capsule network performance for MNIST dataset on highly overlapping digits could work better than CNNs.
In computer vision, a capsule network is a neural network that tries to work inverse graphics. In a sense, the approach tries to reverse-engineer the physical process that produces an image of the world \citep{invers2019}.

The capsule network is composed of many capsules that act like a function, and try to predict the instantiation parameters and presence of a particular object at a given location. 

One key feature of capsule networks is equivariance, which aims at keeping detailed information about the location of the object and its pose throughout the network. For example, if someone rotates the image slightly, the activation vectors also change slightly \citep{Aurelien:2017}. 
One of the limitations of a regular CNN is losing the precise location and pose of the objects in an image. Although this is not a challenging issue when classifying the whole image, it can be a bottleneck for image segmentation or object detection that needs precise location and pose. A capsule, however, can overcome this shortcoming in such applications \citep{Aurelien:2017}. 

Capsule networks have recently received significant attention. This model aims at improving CNNs and RNNs by adding the following capabilities to each source, and target node: (1) the source node has the capability of deciding about the number of messages to transfer to target nodes, and (2) the target node has the capability of deciding about the number of messages that may be received from different source nodes \citep{gong2018information}.

After the success of capsule networks in computer vision tasks \citep{Afshar:2018,mobiny:2018,kumar:2018}, capsule networks have been used in different NLP tasks, including text classification \citep{ kim:2019,Zhao:2018}, multi-label text classification \citep{aly:2019}, sentiment analysis \citep{Wang:2018,Gong:2018}, identifying aggression and toxicity in comments \citep {srivastava:2018}, and zero-shot user intent detection \citep{ xia2018zero}.

\begin{figure}[H]
  \centering
  \includegraphics[scale=0.5]{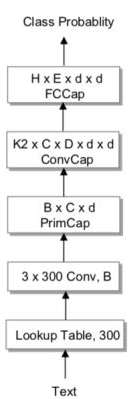}
\caption{The architecture of capsule network proposed by \citet{yang:2018} for text classification}
\label{fig:Yang}
\end{figure}

In capsule networks, the features that are extracted from the text are encapsulated into capsules (groups of neurons).
The first work that applied capsule networks for text classification was done by \citet {yang:2018}. In their research, the performance of the capsule network as a text classification network was evaluated for the first time. Their capsule network architecture includes a standard convolutional layer called n-gram convolutional layer that works as a feature extractor. The second layer is a layer that maps scalar-valued features into a capsule representation and is called the primary capsule layer. The outputs of these capsules are fed to a convolutional capsule layer. In this layer, each capsule is only connected to a local region in the layer below. In the last step, the output of the previous layer is flattened and fed through a feed-forward capsule layer. For this layer, every capsule of the output is considered as a particular class.
 In this architecture, a max-margin loss is used for training the model. Figure \ref{fig:Yang} shows the architecture proposed by \citet{yang:2018}.

Some characteristics of capsules make them suitable for presenting a sentence or document as a vector for text classification. These characteristics include representing attributes of partial entities and expressing semantic meaning in a wide space \citep{ kim:2019}.

\begin{figure}[H]
  \centering
  \includegraphics[scale=0.35]{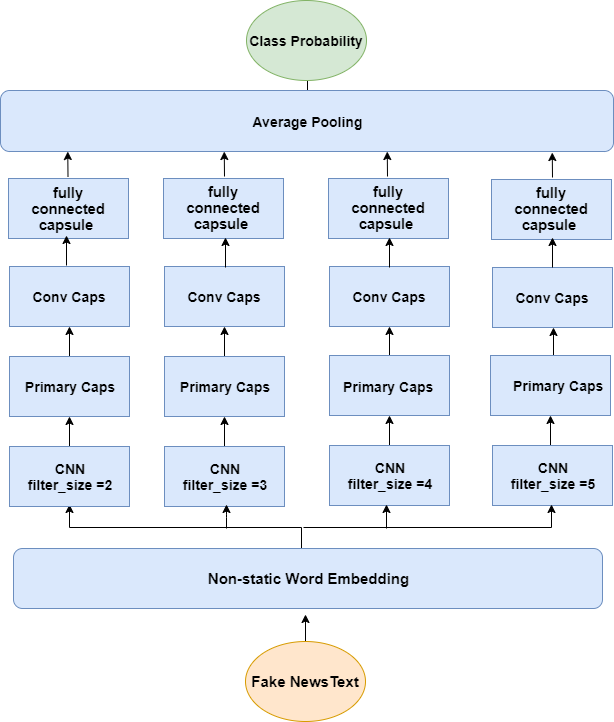}
\caption{The architecture of the proposed non-static capsule network for detecting fake news in medium or long news statements.}
\label{fig:CAPSARCH}
\end{figure}

For fake news identification with different length of statements, our model benefits from several parallel capsule networks and uses average pooling in the last stage. With this architecture, the models can learn more meaningful and extensive text representations on different n-gram levels according to the length of texts.\\
Depending on the length of the news statements, we use two different architectures. Figure \ref{fig:CAPSARCH} depicts the structure of the proposed model for medium or long news statements.  In the model, a non-static word embedding is used as an embedding layer. In this layer, we use 'glove.6B.300d' as a pre-trained word embedding, and use four parallel networks by considering four different filter sizes  2,3,4,5 as n-gram convolutional layers for feature extraction. In the next layers, for each parallel network, there is a primary capsule layer and a convolutional capsule layer, respectively, as presented in Figure \ref{fig:Yang}. A fully connected capsule layer is used in the last layer for each parallel network. At the end, the average polling is added for producing the final result.

\begin{figure}[H]
  \centering
  \includegraphics[scale=0.35]{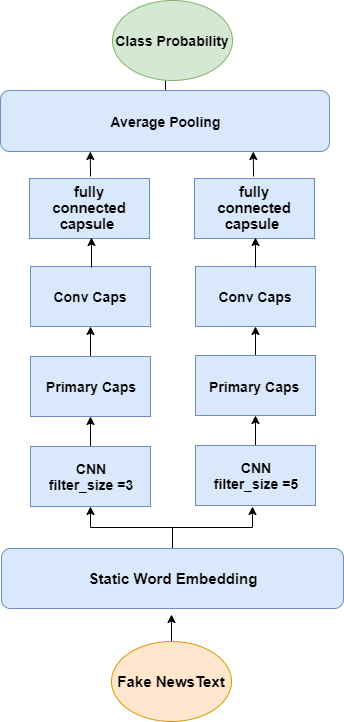}
\caption{The architecture of the proposed static capsule network for detecting fake news in short news statements.}
\label{fig:CAPSARCH3}
\end{figure}

For short news statements, due to the limitation of word sequences, a different structure has been proposed. The layers are like the first model, but only two parallel networks are considered with 3 and 5 filter sizes. In this model, a static word embedding is used. Figure \ref{fig:CAPSARCH3} shows the structure of the proposed model for short news statements.

\section{Evaluation}
\label{sec:Section 4}
\subsection{Dataset}
Several datasets have been introduced for fake news detection. One of the main requirements for using neural architectures is having a large dataset to train the model. In this paper, we use two datasets, namely ISOT fake news \citep{ahmed:2017} and LIAR \citep{Wang:2017}, which have a large number of documents for training deep models. The length of news statements for ISOT is medium or long, and LIAR is short. 

\subsubsection{The ISOT fake news dataset}

 In 2017, \citet{ahmed:2017} introduced a new dataset that was collected from real-world sources\footnote{\url{http://www.uvic.ca/engineering/ece/isot/datasets/index.php}}. This dataset consists of news articles from \texttt{Reuters.com} and \texttt{Kaggle.com} for real news and fake news, respectively. Every instance in the dataset is longer than 200 characters. For each article, the following metadata is available:
article type, article text, article title, article date, and article label (fake or real).
Table \ref{tab:ISOTDataStatistics} shows the type and size of the articles for the real and fake categories.

\begin{table}[H]
	\centering
\begin{tabular}{|c|c|c|c|}
\hline
\multirow{2}{*}{\textbf{News Type}} & \multirow{2}{*}{\textbf{Total size}} & \multicolumn{2}{c|}{\textbf{Subject}} \\ \cline{3-4} 
                                    &                                      & \textbf{Type}      & \textbf{Size}    \\ \hline
\multirow{2}{*}{\textbf{Real-News}} & \multirow{2}{*}{21417}               & World-News         & 10145            \\ \cline{3-4} 
                                    &                                      & Politics-News      & 11272            \\ \hline
\multirow{6}{*}{\textbf{Fake-News}} & \multirow{6}{*}{23481}               & Government-News    & 1570             \\ \cline{3-4} 
                                    &                                      & Middle-east        & 778              \\ \cline{3-4} 
                                    &                                      & US News            & 783              \\ \cline{3-4} 
                                    &                                      & Left-News          & 4459             \\ \cline{3-4} 
                                    &                                      & Politics           & 6841             \\ \cline{3-4} 
                                    &                                      & News               & 9050             \\ \hline
\end{tabular}
\caption{Type and size of every articles per category for ISOT Dataset provided by \citet{ahmed:2017}}
	\label{tab:ISOTDataStatistics}
\end{table}

\subsubsection{The LIAR dataset}
\label{sec:Section 4.1.2}
As mentioned in Section \ref{sec:Section 2}, one of the recent well-known datasets, is provided by \citet{Wang:2017}. 
\citet{Wang:2017} introduced a new large dataset called LIAR, which includes 12.8K human-labeled short statements from \texttt{POLITIFACT.COM} API. Each statement is evaluated by \texttt{POLITIFACT.COM} editor for its validity. Six fine-grained labels are considered for the degree of truthfulness, including pants-fire, false, barely-true, half-true, mostly-true, and true.
The distribution of labels in this dataset are as follows: 1,050 pants-fire labels and a range of 2,063 to 2,638 for other labels.\\
 In addition to news statements, this dataset consists of several metadata as speaker profiles for each news item. These metadata include valuable information about the subject, speaker, job, state, party, and total credit history count of the speaker of the news.
The total credit history count, including the barely-true counts, false counts, half-true counts, mostly-true counts, and pants-fire counts. The statistics of LIAR dataset are shown in Table \ref{tab:LiarDataStatistics}. 
Some excerpt samples from the LIAR dataset are presented in Table \ref{tab:LiarDataSample}.

\begin{table}[H]
	\centering
	\begin{tabular}{|lr|}
		\hline
		\multicolumn{2}{|l|}{LIAR Dataset Statistics} \\\hline
		Training set size & 10,269\\
		Validation set size & 1,284\\
		Testing set size & 1,283\\
		Avg. statement length (tokens) & 17.9 \\\hline 
		\multicolumn{2}{|l|}{Top-3 Speaker Affiliations}\\ 
		Democrats & 4,150\\
		Republicans & 5,687\\
		None (e.g., FB posts) & 2,185\\\hline
		
	\end{tabular}
	\caption{The LIAR dataset statistics provided by \citet{Wang:2017}}
	\label{tab:LiarDataStatistics}
\end{table}

\begin{table}[H]
	\centering
	\begin{tabular}{|p{0.35\linewidth}|p{0.35\linewidth}|p{0.35\linewidth}|}
		\hline
		Sample 1 & Sample 2 & Sample 3 \\\hline
		\begin{scriptsize}
		 \textbf{Statement:} “The last quarter, it was just
		 announced, our gross domestic product
		 was below zero. Who ever heard of this?
		 It's never below zero.”

		 \textbf{Speaker:} Donald Trump

		 \textbf{Context:} presidential announcement speech

		 \textbf{Label:} Pants-fire

		 \textbf{Justification:} According to Bureau of
		 Economic Analysis and the National Bureau
		 of Economic Research, the growth
		 in the gross domestic product has been
		 below zero 42 times over 68 years. That's
		 a lot more than “never.” We rate his
		 claim Pants on Fire!
		 		\end{scriptsize}
		 & 
		 		\begin{scriptsize}
		 \textbf{Statement:} “Newly Elected Republican
		 Senators Sign Pledge to Eliminate Food
		 Stamp Program in 2015.”

		 \textbf{Speaker:} Facebook posts

		 \textbf{Context:} social media posting

		 \textbf{Label:} Pants-fire

		 \textbf{Justification:} More than 115,000 social
		 media users passed along a story
		 headlined, “Newly Elected Republican
		 Senators Sign Pledge to Eliminate Food
		 Stamp Program in 2015.” But they failed
		 to do due diligence and were snookered,
		 since the story came from a publication
		 that bills itself (quietly) as a
		 “satirical, parody website.” We rate the
		 claim Pants on Fire. 
		 		 		\end{scriptsize}
		 &
		 		\begin{scriptsize}
		 \textbf{Statement:} “Under the health care law,
		 everybody will have lower rates, better
		 quality care and better access.”

		 \textbf{Speaker:} Nancy Pelosi
		 Context: on ’Meet the Press’

		 \textbf{Label:} False

		 \textbf{Justification:} Even the study which
		 Pelosi’s staff cited as a source of that
		 the statement suggested that some people
		 would pay more for health insurance.
		 Analysis at the state level found the
		 same thing. The general understanding
		 of the word “everybody” is every person.
		 The predictions do not back that up.
		 We rule this statement False.
		 		 		\end{scriptsize}
		  \\\hline
	\end{tabular}
	\caption{Three random excerpts from the LIAR dataset.}
	\label{tab:LiarDataSample}
\end{table}

\subsection{Experimental setup}
The experiments of this paper were conducted on a PC with Intel Core i7 6700k, 3.40GHz CPU; 16GB RAM; Nvidia GeForce GTX 1080Ti GPU in a Linux workstation. For implementing the proposed model, the Keras library \citep{chollet2015keras} was used, which is a high-level neural network API.

\subsection{Evaluation metrics}
The evaluation metric in our experiments is the classification accuracy. Accuracy is  the ratio of correct predictions to the total number of samples and is computed as:

\begin{align}
 Accuracy = \frac{TP + TN}{TP + TN + FN + FP}
 \label{eq:1}
 \end{align}
 \\
Where TP is represents the number of True Positive results, FP represents the number of False Positive results, TN represents the number of True Negative results, and FN represents the number of False Negative results.

\section{Results}
\label{sec:Section 5}
For evaluating the effectiveness of the proposed model, a series of experiments on two datasets were performed. These experiments are explained in this section and the results are compared to other baseline methods. We also discuss the results for every dataset separately.

\subsection{Classification for ISOT dataset}
\label{sec:Section 5.2}
As mentioned in Section \ref{sec:Section 4},  \citet{ahmed:2017} presented the ISOT dataset. According to the baseline paper, we consider 1000 articles for every set of real and fake articles, a total of 2000 articles for the test set, and the model is trained with the rest of the data.

First, the proposed model is evaluated with different word embeddings that described in Section \ref{sec:non-static}. Table  \ref{tab:Table_ISOT1} shows the result of applying different word embeddings for the proposed model on ISOT, which consists of medium and long length news statements. The best result is achieved by applying the non-static embedding.

\begin{table}[H]
  \centering
  \begin{tabular}{|c|c|c|c|}
    \hline
    Model &static& non-static accuracy& multi-channel accuracy  \\\hline
    Proposed model &  99.6 & 99.8 & 99.1 \\\hline
  \end{tabular}
  \caption{Result of proposed model with different word embedding models}\label{tab:Table_ISOT1}
\end{table}

\citet{ahmed:2017} evaluated different machine learning methods for fake news detection on the ISOT dataset, including the Support Vector Machine (SVM), the Linear Support Vector Machine (LSVM), the K-Nearest Neighbor (KNN), the Decision Tree (DT), the Stochastic Gradient Descent (SGD), and the Logistic regression (LR) methods.

Table \ref{tab:Table_ISOT} shows the performance of non-static capsule network for fake news detection in comparison to other methods. The accuracy of our model is 7.8\% higher than the best result achieved by LSVM.

\begin{table}[H]
  \centering
  \begin{tabular}{|c|c|c|}
    \hline
    Model &Meta-data& Test Accuracy  \\\hline
    SVM & Article Text & 0.86 \\\hline
    LSVM & Article Text & 0.92 \\\hline
    KNN & Article Text & 0.83 \\\hline
    Decision Tree & Article Text & 0.89 \\\hline
    SGD & Article Text & 0.89 \\\hline
    Linear Regression & Article Text & 0.89 \\\hline
    proposed non-static capsule network & Article Text & \textbf{0.998} \\\hline

  \end{tabular}
  \caption{Comparison of non-static capsule network result with Result of \citet{ahmed:2017}}\label{tab:Table_ISOT}
\end{table}

\subsection{Discussion}
\label{sec:Section 5.3}
The proposed model can predict true labels with high accuracy reaching in a very small number of wrong predictions. Table \ref{tab:Table_6} shows the titles of two wrongly predicted samples for detecting fake news. To have an analysis on our results, we investigate the effects of sample words that are represented in training statements that tagged as real and fake separately. 

\begin{table}[H]
  \centering
\begin{tabular}{|c|c|c|}
\hline
\textbf{Statement title}                                                                                                             & \textbf{Predicted}    & \textbf{True label}   \\ \hline
\multirow{2}{*}{\begin{tabular}[c]{@{}c@{}}Factbox: In U.S. Senate, \\ Democrats represent highest-tax states\end{tabular}}          & \multirow{2}{*}{fake} & \multirow{2}{*}{real} \\
                                                                                                                                     &                       &                       \\ \hline
\multirow{2}{*}{\begin{tabular}[c]{@{}c@{}}Trump Vows To Save America From \\ ‘Curse’ Of Functional Health Care System\end{tabular}} & \multirow{2}{*}{real} & \multirow{2}{*}{fake} \\
                                                                                                                                     &                       &                       \\ \hline
\end{tabular}
  \caption{Two samples with wrong prediction}\label{tab:Table_6}
\end{table}
\begin{table}[H]
  \centering
\begin{tabular}{|c|c|c|}
\hline
Data                                                                                      & \multicolumn{1}{l|}{\textbf{Word tokens}} & \multicolumn{1}{l|}{\textbf{Word types}} \\ \hline
\textbf{\begin{tabular}[c]{@{}c@{}}Training data\\ With real label\end{tabular}}          & 8264220                                  & 76213                                   \\ \hline
\textbf{\begin{tabular}[c]{@{}c@{}}Training data\\ With fake label\end{tabular}}         & 10115367                                 & 92613                                   \\ \hline
\textbf{\begin{tabular}[c]{@{}c@{}}Sample 1\\ (Predicted fake but is real)\end{tabular}} & 257                                      & 162                                     \\ \hline
\textbf{\begin{tabular}[c]{@{}c@{}}Sample 2\\ (Predicted real but is fake)\end{tabular}} & 413                                      & 243                                     \\ \hline
\end{tabular}
  \caption{The number of word tokens and word types of training data and samples}\label{tab:Table_7}
\end{table}

 For this work, all of the words and their frequencies are extracted from the two wrong samples and both real and fake labels of the training data. Table \ref{tab:Table_7} shows the information of this data. Then for every wrongly predicted sample, stop-words are omitted, and words with a frequency of more than two are listed. After that, all of these words and their frequency in real and fake training datasets are extracted. In this part, the frequencies of these words are normalized. Table \ref{tab:Table_8} and Table \ref{tab:Table_9} show the normalized frequencies of words for each sample respectably. In these tables, for ease of comparison, the normalized frequencies of real and fake labels of training data and the normalized frequency for each word in every wrong sample are multiplied by 10. 
 
The label of Sample 1 is predicted as fake, but it is real. In Table \ref{tab:Table_8}, six most frequent words of Sample 1 are listed, the word \textit{"tax"} is presented 2 times more than each of the other words in Sample 1,   and this word in the training data with real labels is obviously more frequent. In addition to this word, for other words like \textit{"state"}, the same observation exists. 
\begin{table}[H]
  \centering
\begin{tabular}{|c|c|c|c|}
\hline
\multirow{2}{*}{\textbf{Word}} & \multicolumn{3}{c|}{\textbf{Normalized frequency}}                                                                                                                                                                                       \\ \cline{2-4} 
                               & \textbf{\begin{tabular}[c]{@{}c@{}}Sample 1\\ (Real label)\end{tabular}} & \textbf{\begin{tabular}[c]{@{}c@{}}Fake label \\ training data\end{tabular}} & \textbf{\begin{tabular}[c]{@{}c@{}}Real label \\ training data\end{tabular}} \\ \hline
\textbf{tax}                   & 0.350195                                                                   & 0.00377643                                                                    & 0.00901355                                                                    \\ \hline
\textbf{would}                 & 0.155642                                                                   & 0.02250141                                                                    & 0.036057849                                                                   \\ \hline
\textbf{deduction}             & 0.116732                                                                   & 0.00002471                                                                    & 0.00011737                                                                    \\ \hline
\textbf{salt}                  & 0.116732                                                                   & 0.00009392                                                                    & 0.000053242                                                                   \\ \hline
\textbf{senate}                & 0.116732                                                                   & 0.00346898                                                                    & 0.01006024                                                                    \\ \hline
\textbf{states}                & 0.116732                                                                   & 0.00977325                                                                    & 0.019226255                                                                   \\ \hline
\end{tabular}
  \caption{Normalized frequency for words in sample 1 and training data with fake and real label}\label{tab:Table_8}
\end{table}
The text of Sample 2 is predicted as real news, but it is fake.  Table \ref{tab:Table_9} lists six frequent words of Sample 2. The two most frequent words of this text are \textit{"trump"} and \textit{"sanders"}. These words are more frequent in training data with fake labels than the training data with real labels. \textit{"All"} and \textit{"even"} are two other frequent words, We use \textit{"even"} to refer to something surprising, unexpected, unusual or extreme\footnote{\url{https://dictionary.cambridge.org/grammar/british-grammar/even}} and \textit{"all"} means every one, the complete number or amount or the whole. \footnote{\url{https://dictionary.cambridge.org/grammar/british-grammar/all}} therefore, a text that includes these words has more potential to classify as a fake news. These experiments show the strong effect of the sample words frequency on the prediction of the labels.
\begin{table}[H]
  \centering
\begin{tabular}{|c|c|c|c|}
\hline
\multirow{2}{*}{\textbf{Word}} & \multicolumn{3}{c|}{\textbf{Normalized frequency}}                                                                                                                                                                                       \\ \cline{2-4} 
                               & \textbf{\begin{tabular}[c]{@{}c@{}}Sample 2\\ (Fake label)\end{tabular}} & \textbf{\begin{tabular}[c]{@{}c@{}}Fake label \\ training data\end{tabular}} & \textbf{\begin{tabular}[c]{@{}c@{}}Real label \\ training data\end{tabular}} \\ \hline
\textbf{trump}                 & 0.193705                                                                   & 0.07903322                                                                    & 0.05449879                                                                    \\ \hline
\textbf{sanders}               & 0.145278                                                                   & 0.00388419                                                                    & 0.00258585                                                                    \\ \hline
\textbf{all}                   & 0.121065                                                                   & 0.02453594                                                                    & 0.01085523                                                                    \\ \hline
\textbf{coverage}              & 0.121065                                                                   & 0.00106472                                                                    & 0.0010975                                                                     \\ \hline
\textbf{donald}                & 0.121065                                                                   & 0.01681699                                                                    & 0.01160908                                                                    \\ \hline
\textbf{even}                  & 0.096852                                                                   & 0.01315128                                                                    & 0.00444204                                                                    \\ \hline
\end{tabular}
  \caption{Normalized frequency for words in sample 2 and training data with fake and real label}\label{tab:Table_9}
\end{table}
\subsection{Classification for the LIAR dataset }

As mentioned in Section \ref{sec:Section 4.1.2}, the LIAR dataset is a multi-label dataset with short news statements. In comparison to the ISOT dataset, the classification task for this dataset is more challenging. We evaluate the proposed model while using different metadata, which is considered as speaker profiles. Table \ref{tab:Table_LIAR} shows the performance of the capsule network for fake news detection by adding every metadata. The best result of the model is achieved by using history as metadata. The results show that this model can perform better than state-of-the-art baselines including hybrid CNN \citep{Wang:2017} and LSTM with attention \citep{long2017fake} by 3.1\% on the validation set and 1\% on the test set. 

\begin{table}[H]
  \centering
\begin{tabular}{|c|c|c|c|l|l|l|l|l|}
\hline
\multirow{3}{*}{\textbf{Model}}                                              & \multicolumn{8}{c|}{\textbf{Metadata}}                                                                                                               \\ \cline{2-9} 
                                                                              & \multicolumn{2}{c|}{\textbf{Party}} & \multicolumn{2}{c|}{\textbf{State}} & \multicolumn{2}{c|}{\textbf{Job}} & \multicolumn{2}{c|}{\textbf{History}} \\ \cline{2-9} 
                                                                              & \textbf{Valid}    & \textbf{Test}   & \textbf{Valid}    & \textbf{Test}   & \textbf{Valid}   & \textbf{Test}  & \textbf{Valid}     & \textbf{Test}    \\ \hline
\textbf{\begin{tabular}[c]{@{}c@{}}Hybrid CNN\\  \citet{Wang:2017}\end{tabular}}    & 25.9              & 24.8            & 24.6              & 25.6            & \textbf{27}      & \textbf{25.8}  & 24.6               & 24.1             \\ \hline
\textbf{\begin{tabular}[c]{@{}c@{}}LSTM attention\\ \citet{long2017fake}\end{tabular}} & 25.3              & \textbf{25.7}   & 26.6              & \textbf{26.8}   & 25.8             & 25.7           & 37.8               & 38.5             \\ \hline
\textbf{\begin{tabular}[c]{@{}c@{}}Proposed model\\ Capsule network\end{tabular}}                                                   & \textbf{26.1}     & 24              & \textbf{27}       & 24.3            & 25.4             & 25.1           & \textbf{40.9}      & \textbf{39.5}    \\ \hline
\end{tabular}
  \caption{Comparison of capsule network result with other baseline}\label{tab:Table_LIAR}
\end{table}
\subsubsection{Discussion}
\begin{figure}[H]
  \centering
  \includegraphics[scale=0.5]{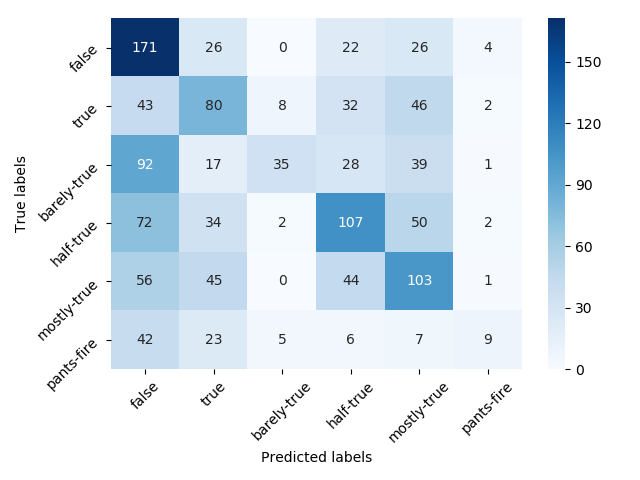}
\caption{Confusion matrix of classification using proposed model for LIAR dataset}
\label{fig:CM_LIAR}
\end{figure}

Figure \ref{fig:CM_LIAR} shows the confusion matrix of the best classification using the proposed model for the test set. The model classifies false, half-true, and mostly-true news with more accuracy. Nevertheless, it is difficult to distinguish between true and mostly-true and also between barely-true and false.  The worst accuracy is for classifying pants-fire. For these labels, detecting the correct label is more challenging, and many pants-fire texts are predicted as false.

\section{Conclusion}
\label{sec:Section 6}
In this paper, we apply capsule networks for fake news detection. We propose two architectures for different lengths of news statements. We apply two strategies to improve the performance of the capsule networks for the task. First, for detecting the medium or long length of news text, we use four parallel capsule networks that each one extracts different n-gram features (2,3,4,5) from the input texts. Second, we use non-static embedding such that the word embedding model is incrementally up-trained and updated in the training phase.\\
 Moreover, as a fake news detector for short news statements, we use only two parallel networks with 3 and 5 filter sizes as a feature extractor and static model for word embedding. For evaluation, two datasets are used. The ISOT dataset as a medium length or long news text and LIAR as a short statement text. The experimental results on these two well-known datasets showed improvement in terms of accuracy by 7.8\% on the ISOT dataset and  3.1\% on the validation set and 1\% on the test set of the LIAR dataset.


\bibliography{myreferences}

\end{document}